\documentclass[10pt]{article} % For LaTeX2e
\usepackage[accepted]{tmlr}
\usepackage{amsmath,amsfonts,bm}

\def\eqref#1{equation~\ref{#1}}
\def\1{\bm{1}}

\DeclareMathAlphabet{\mathsfit}{\encodingdefault}{\sfdefault}{m}{sl}
\SetMathAlphabet{\mathsfit}{bold}{\encodingdefault}{\sfdefault}{bx}{n}

\newcommand{\E}{\mathbb{E}}

\newcommand{\R}{\mathbb{R}}

\DeclareMathOperator*{\argmin}{arg\,min}

\usepackage{hyperref}
\usepackage{url}
\usepackage{amsthm} 
\usepackage{wrapfig}
\usepackage{graphicx} 
\usepackage{amssymb, amsmath}

\def\bN{{\mathbb N}}

\def\bR{{\mathbb R}}

\def\bE{{\mathbb E}}

\def\X{{\mathcal{X}}}
\def\Y{{\mathcal{Y}}}
\def\C{{\mathcal{C}}}
\def\W{{\mathcal{W}}}

\def\F{{\mathcal{F}}}
\def\H{{\mathcal{H}}}

\def\E{{\mathcal{E}}}

\def\P{{\mathcal{P}}}
\def\D{{\mathcal{D}}}
\def\R{{\mathcal{R}}}

\def\I{{\mathcal{I}}}
\def\E{{\mathcal{E}}}

\def\e{{\varepsilon}}

\newcommand{\kf}[1]{\textcolor{black}{#1}}
\newcommand{\st}[1]{\textcolor{black}{#1}}
\newcommand{\kff}[1]{\textcolor{black}{#1}}

\newtheorem{theo}{Theorem}

\newtheorem{lemm}[theo]{Lemma}%
{\tiny }%

\usepackage{setspace}
\usepackage {diagbox}

\title{Out-of-Distribution Optimality of Invariant Risk Minimization }

\author{\name Shoji Toyota \email shoji@ism.ac.jp \\
      \addr The Institute of Statistical Mathematics
      \AND
     \name Kenji Fukumizu \email fukumizu@ism.ac.jp \\
      \addr The Institute of Statistical Mathematics
      }

\def\month{01}  % Insert correct month for camera-ready version
\def\year{2024} % Insert correct year for camera-ready version
\def\openreview{\url{https://openreview.net/forum?id=pWsfWDnJDa}} % Insert correct link to OpenReview for camera-ready version

\begin{document}

\maketitle

\begin{abstract}
Deep Neural Networks often inherit spurious correlations embedded in training data and hence may fail to generalize to unseen domains, which have different distributions from the domain to provide training data. \citet{arjovsky2020} introduced the concept {\em out-of-distribution (o.o.d.) risk}, which is the maximum risk among all domains, and formulated the issue caused by spurious correlations as a minimization problem of the o.o.d.~risk. Invariant Risk Minimization (IRM) is considered to be a promising approach to minimize the o.o.d.~risk: IRM estimates a minimum of the o.o.d.~risk by solving a bi-level optimization problem. 
While IRM has attracted considerable attention with empirical success,  it comes with few theoretical guarantees. Especially, a solid theoretical guarantee that the bi-level optimization problem gives the minimum of the o.o.d.~risk has not yet been established.  Aiming at providing a theoretical justification for IRM, this paper rigorously proves  that a solution to the bi-level optimization problem minimizes the o.o.d.~risk under certain conditions. The result also provides sufficient conditions on distributions providing training data and on a dimension of a feature space for the bi-leveled optimization problem to minimize  the o.o.d.~risk.

\end{abstract}

%-------------------------------------------------------------------------------------------------------------------
\section{Introduction}\label{sec:intro}
%-------------------------------------------------------------------------------------------------------------------

%Training data used in machine learning may contain features that are spuriously correlated to the response variables of data.  
Training data used in supervised learning may contain features that are spuriously correlated to the response variables of data.  Deep Neural Networks (DNNs) often learn such spurious correlations embedded in the data and hence may fail to predict desirable response variables of test data generated by a distribution that is different from the one to  provide training data. To list a few examples, in a classification of animal images, models obtained by conventional procedures tend to misclassify cows on sandy beaches because most training pictures are captured in green pastures and DNNs inherit context information in training \citep{beery2018, zotero-64}. Another example is learning from medical data. Systems trained with data collected in one hospital do not generalize well to other hospitals; DNNs unintentionally extract environmental factors specific to a particular hospital in training \citep{albadawy2018,perone2019,zotero-107}.

\citet{arjovsky2020} introduced the concept {\em out-of-distribution (o.o.d.) risk} to formulate the issue caused by spurious correlations. 
Let $\X$ and $\Y$ be measurable spaces of explanatory and response variables respectively. Let $\E$ be a set with each element $e \in \E$ called the {\em domain} (or environment) $e$. Assume that for a given domain $e \in \E$, there corresponds  a corresponding random variable $(X^e, Y^e)$ that takes values in $\X \times \Y$ with its probability law $P_{X^e, Y^e}$.  Assume we are given training datasets $\D^e:= \{(x^e_i, y^e_i) \}_{i=1}^{n^e} \sim P_{X^e, Y^e}$ i.i.d.~from multiple domains $ \E_{tr} \subset \E$. For a given predictor $f :\X \rightarrow \Y$, $$ \R^e (f) := \int l(f(x), y) dP_{X^e, Y^e}$$ denotes  the risk of $f$ on domain $e$.  
The o.o.d.~risk of the predictor $f$ is as follows:
 \begin{equation}\label{eq:OOD-Bayes rule}
  \R^{o.o.d.}(f) := \max_{e \in \E} \R^e(f), 
 \end{equation} which is the worst-case risk over $\E$ including unseen domain $\E - \E_{tr}$. \citet{arjovsky2020} formulated the problem caused by spurious correlations as a minimization problem of the o.o.d. risk (\ref{eq:OOD-Bayes rule}):
 \begin{equation}\label{eq:oodMinimization}
     \min_{f \in \F} \R^{o.o.d.} (f),
 \end{equation}
 where $\F$ is the set of all measurable functions $f: \X \rightarrow \Y$.

It is difficult to directly solve the o.o.d.~risk minimization (\ref{eq:oodMinimization}) since we can not evaluate the maximum of risks among all domains $\E$,  including unseen domains $\E - \E_{tr}$, only by data from training domains $\E_{tr} \subset \E$. 
Invariant Risk Minimization (IRM) is a rapidly developing approach to  the challenging o.o.d.~risk minimization \citep{arjovsky2020}. Its proposed predictor $f:= w \circ \Phi $ is composed of two maps: a feature map $\Phi: \X \rightarrow \H$, which is  called an {\em invariance},  
and a predictor  $w : \H \rightarrow \Y$ which estimates the response variable of feature $\Phi(x)$. Here, for a given feature space $\H$, we call a measurable function $\Phi:  \X \rightarrow \H$ an invariance when it holds that  $P_{Y^{e_1} | \Phi(X^{e_1})} = P_{Y^{e_2} | \Phi(X^{e_2})}\text{ for any }e_1, e_2 \in \E\footnote{The definition is based on conditional independence \citep{peters2016, koyama2021, rojas-carulla2018a}, while \citet{arjovsky2020, ahuja2020} used a different type of invariances based on  $\argmin \nolimits_{w}  \R^e(w\circ \Phi)$ instead of $P_{Y^e | \Phi(X^e)}$.  Throughout the paper, we argue by adopting the definition based on conditional independence.}. $
\citet{arjovsky2020} estimated the two maps  by solving the bi-leveled optimization problem
\begin{equation}\label{eq:IRM}
 \min \nolimits_{\Phi \in \I_{tr}, \\
w \in \W}  \sum_{e \in \E_{tr}} \R^{e}(w \circ \Phi ),
\end{equation}
where $\W$ is a model of predictors $w: \H \rightarrow \Y$ and $\I_{tr}$ is the set of invariances captured by training domains $ \E_{tr}$:
\begin{equation}\label{inv_trdom}
\I_{tr} :=\bigl\{ \Phi : \X \rightarrow \H \left| \right. \bigl. P_{Y^{e_1} | \Phi(X^{e_1})} = P_{Y^{e_2} | \Phi(X^{e_2})}\text{ for any }e_1, e_2 \in \E_{tr}  \bigr\}.
\end{equation}
Influenced by the seminal study, several alternative bi-leveled optimization problems have been proposed \citep{ahuja2020, chang2020, ahuja2021, ahuja2021a, lin2022, zhou2022, liu2021, liu2021a, lu2022, koyama2021, parascandolo2022, krueger2021, toyota2022, lin2022b, huh2022, rame2022, pogodin2023, chen2023, tam2023}. For example, \citet{ahuja2020} proposed a novel bi-leveled optimization problem leveraging the principles of game theory. The recently proposed Maximal Invariant Predictor \citep{koyama2021} employed a new bi-leveled problem grounded in the concept of information theory.

  While IRM is widely recognized as a promising approach for  the o.o.d.~risk minimization (\ref{eq:oodMinimization}), it comes with few theoretical guarantees; especially, a mathematical guarantee that the bi-level optimization problem (\ref{eq:IRM}) gives the minimum of the o.o.d.~risk (\ref{eq:OOD-Bayes rule}) has not yet been established.\footnote{\st{Since it is  difficult to solve the bi-leveled optimization problem (\ref{eq:IRM}), several papers have proposed optimization methods for (\ref{eq:IRM}) such as IRMv1 \citep{arjovsky2020} or  Invariant Rationalization \citep{chang2020}. While their optimization ability for solving (\ref{eq:IRM}) should also be discussed theoretically, this paper does not address it and only focuses on the problem of whether, assuming that  (\ref{eq:IRM}) can be solved completely, the resulting predictor minimizes the o.o.d.~risk.}} The original IRM paper did not mention any theoretical properties for the minimum of  (\ref{eq:IRM}). \citet{rosenfeld2021} proved that, assuming that data follow a simple linear Gaussian structural equation model (SEM),  a predictor obtained by  (\ref{eq:IRM}) makes a prediction relying only on a feature of $X^e \in \X$ whose distribution does not depend on domains \citep[Section 5]{rosenfeld2021}. However, their analysis did not focus on relations between the bi-level optimization problem (\ref{eq:IRM}) and the  o.o.d.~risk  (\ref{eq:oodMinimization}). More recently, \citet{kamath2021} provided an example of distributions on which a  minimum  of (\ref{eq:IRM}) does not minimize the o.o.d~risk \citep[Section 4]{kamath2021}.  However, their analysis assumed that data follow  particular SEMs constructed to derive the case where (\ref{eq:IRM}) does not provide a minimum of the o.o.d.~risk; for verifying the o.o.d.~performance of the bi-leveled optimization problem (\ref{eq:IRM}),  it should be analyzed under more general assumptions on distributions.

Aiming at providing a theoretical justification for IRM,  this paper rigorously proves that a solution to the bi-leveled optimization problem (\ref{eq:IRM}) also minimizes the o.o.d.~risk (\ref{eq:OOD-Bayes rule}); formally speaking, we prove that the inclusion 
\begin{equation}\label{eq:IRMoptimal}
 \argmin \nolimits_{\Phi \in \I_{tr}, \\
w \in \W}  \sum_{e \in \E_{tr}} \R^{e}(w \circ \Phi ) \subset \argmin_{f \in \F}  \R^{o.o.d.}(f ) 
\end{equation} is attained under certain conditions. The result also provides sufficient conditions on the training domains $\E_{tr}$ and the feature space $\H$ to minimize  the o.o.d~risk.  
In our analysis, we set distributions on domains $\E$ by the ones proposed in \citet{rojas-carulla2018a}.  The distributions do not rely on  any specific SEM structures, unlike existing theoretical analysis of  IRM  \citep{rosenfeld2021, kamath2021}, %and have been  used for the analysis of methods related to  invariances \citep{rojas-carulla2018a,  toyota2022}.
and they are  used for the analysis of methods related to  invariances \citep{rojas-carulla2018a,  toyota2022}.

The rest of the paper is organized as follows.  Section \ref{sec:Main}  illustrates  two main theorems. Section \ref{subsec:theo_in_MSE} provides the first main theorem, which states that the inclusion (\ref{eq:IRMoptimal}) is achieved  in the regression case.  In Section \ref{subsec:theo_in_CE}, we extend the first theorem to the classification case.  The novelty and significance of these two theorems are discussed in Section \ref{subsec:nov_and_sig}. We provide a review of the prior works concerning the relationship between the bi-leveled optimization problem (\ref{eq:IRM}) and the o.o.d.~risk (\ref{eq:OOD-Bayes rule}) in Section \ref{sec:related}. The two main theorems stated in Section \ref{sec:Main}  are proved in  Section \ref{sec:proof}. Section \ref{sec:conclusion} is devoted to brief concluding remarks.

%The remainder of the paper is organized as follows.  Section \ref{sec:Main}  illustrates  two main theorems. Section \ref{subsec:theo_in_MSE} provides the first main theorem, which states that the inclusion (\ref{eq:IRMoptimal}) is achieved  in the regression case.  In Section \ref{subsec:theo_in_CE}, we extend the first theorem to the classification case.  Novelties and significance in these two theorems are discussed in Section \ref{subsec:nov_and_sig}. We provide a review of the prior works concerning the relationship between the bi-leveled optimization problem (\ref{eq:IRM}) and the o.o.d.~risk (\ref{eq:OOD-Bayes rule}) in Section \ref{sec:related}. The two main theorems stated in Section \ref{sec:Main}  are proved in  Section \ref{sec:proof}. Section \ref{sec:conclusion} is devoted to brief concluding remarks.

%-------------------------------------------------------------------------------------------------------------------------------------------------------------------

%-------------------------------------------------------------------------------------------------------------------------------------------------------------------
\section{Main Results}\label{sec:Main}
%-------------------------------------------------------------------------------------------------------------------------------------------------------------------

We explain the settings and assumptions  persisting throughout our analysis.

We set  domains $\{ (X^e, Y^e) \}_{e \in \E}$ by the ones proposed in \citet{rojas-carulla2018a}. Let $\X:= \X_1 \times\X_2$ where $\X_1 := \bR^{d_1}$ and $\X_2 := \bR^{d_2}$ with $d_1,d_2 \in \bN_{>0}$,  and $(X_1^I, Y^I)$ be a fixed random variable on $\X_1 \times \Y$. \citet{rojas-carulla2018a}  defined the domain set $\E$ by  all the probability distributions with the fixed conditional distribution $P_{  Y^I|X^I_1 }$; namely,  denoting $\Phi^{\X_1} : \X \rightarrow \X_1$ a projection onto $\X_1$,  $\{ (X^e, Y^e )  \}_{e \in \E}$ is defined by 
\begin{equation}\label{theo:assump1}
\{ (X^e, Y^e )  \}_{e \in \E} := \left\{ (X, Y): \text{a random variable on }  \X \times \Y \left| P_{Y| \Phi^{\X_1}(X)} = P_{  Y^I|X^I_1 } \right. \right\}.
\end{equation}
Note that, under the setting (\ref{theo:assump1}), the projection $\Phi^{\X_1}: \X \rightarrow \X_1$ is an invariance among $\E$, because $P_{Y | \Phi^{\X_1}(X)} = P_{  Y^I|X^I_1 }$ for any $(X, Y)\in \{ (X^e, Y^e) \}_{e \in \E}$. For simplicity of theoretical analysis, we assume that the  conditional distribution $P_{  Y^I|X^I_1 }$ has a probability density function $p^I(y|x_1)$.

We explain assumptions about the feature space and models. The feature space $\H$ for an invariance $\Phi \in \I_{tr}$ is assumed to be the multi-dimensional Euclidean space $\bR^{d_\H}$.  Moreover, we assume that $\Phi   \in \I_{tr}$ and $w \in \W$ in the minimization problem (\ref{eq:IRM}) run only continuous functions; namely, we investigate the property of a solution for   
\begin{equation}\label{eq:IRM_conti}
 \min \nolimits_{\Phi \in \I_{tr}^{\C_0}, \\
w \in \W^{\C_0}}  \sum_{e \in \E_{tr}} \R^{e}(w \circ \Phi ),
\end{equation}
where $\W^{\C_0}$ is the set of all {\em continuous} functions $w: \H \rightarrow \Y$ , and $\I_{tr}^{\C_0}$ is the set of {\em continuous} invariances captured by a training domain $ \E_{tr}$:
$$\I_{tr}^{\C_0} :=\bigl\{ \Phi : \X \rightarrow \H \left| \right. \bigl. P_{Y^{e_1} | \Phi(X^{e_1})} = P_{Y^{e_2} | \Phi(X^{e_2})}\text{ for any }e_1, e_2 \in \E_{tr},~~\Phi: \text{continuous } \bigr\}. $$

%-------------------------------------------------------------------------------------------------------------------------------------------------------------------
\subsection{Case I: Least Square Loss}\label{subsec:theo_in_MSE}
%-------------------------------------------------------------------------------------------------------------------------------------------------------------------

First, we consider the case where $\Y = \bR^{d_\Y}$ ($d_\Y \in \bN_{>0}$) and $l$ is the least square loss; that is, for a given predictor $f: \X \rightarrow \Y$,  its risk $\R^e (f)$ on $(X^e, Y^e) \in \{ (X^e, Y^e )  \}_{e \in \E}$ is given by
$$  \mathcal{R}^e (f) := \int \| y - f(x) \|^2 dP_{X^e, Y^e}. $$

The following theorem ensures that the optimization problem (\ref{eq:IRM_conti}) provides a solution for the o.o.d.~risk minimization problem (\ref{eq:oodMinimization}) under four conditions:
\begin{theo}[o.o.d.~optimality of the bi-leveled optimization problem (\ref{eq:IRM_conti}) under least square loss setting]\label{theo:LSLsetting}
Domains $\{ (X^e, Y^e)\}_{e \in \E}$ are assumed to be (\ref{theo:assump1}). We also assume that the following four conditions hold:
\begin{itemize}
    \item[(i)] $\I_{tr}^{C_0} = \I^{C_0}$, where $\I^{\C_0}$ is the set of continuous invariances captured by all domains $ \E$, not training domains $ \E_{tr}$:
    $$\I^{\C_0} :=\bigl\{ \Phi : \X \rightarrow \H \left| \right. \bigl. P_{Y^{e_1} | \Phi(X^{e_1})} = P_{Y^{e_2} | \Phi(X^{e_2})}\text{ for any }e_1, e_2 \in \E,~~\Phi: \text{continuous } \bigr\}. $$ 
    \item[(ii)]  $\bigcup_{e \in \E_{\text{tr}}}\text{supp}(P_{\Phi^{\X_1}(X^e)}) = \X_1$. Here, for probability measure $\mu$ on $\X_1$, $\text{supp}(\mu)$ is defined by $$\text{supp}(\mu) := \overline{ \left\{x_1 \in \X_1 \left|  N_{x_1}: \st{\text{open neighborhood around }x_1}  \Rightarrow \mu(N_{x_1}) > 0 \right. \right\}}. $$ %\citep{parthasarathy2005} 
    %\footnote{The definition is based on  \citet{parthasarathy2005} (Definition 2.1).}$$
    \item[(iii)] \st{The dimensions $d_1$ and $d_\H$ on the subspace $\X_1\subset \X$ of the input space $\X$ and the feature space $\H = \bR^{d_\H}$ satisfy $d_1 \leq d_\H$.}
    \item[(iv)] $P_{Y^I | X_1^I}$ has a continuous probability density function $p^I(y | x_1)$. Here, we call $p^I(y | x_1)$ continuous when correspondence $\X_1 \times \Y \in (x_1, y) \longmapsto p^{I} (y | x_1)$ is continuous.
\end{itemize}

Then, we have
\begin{equation}\label{eq:IRMoptimal_LSE}
 \argmin \nolimits_{\Phi \in \I_{tr}^{\C_0}, \\
w \in \W^{\C_0}}  \sum_{e \in \E_{tr}} \R^{e}(w \circ \Phi ) \subset \argmin_{f \in \F}  \R^{o.o.d.}(f ).
\end{equation}
 Here, $\F$ is the set of all measurable functions \st{$f: \X \rightarrow \Y$}. 
\end{theo}

We explain the feasibilities and interpretations of the above four conditions.

\paragraph{Condition (i): } Condition (i) implies that invariances captured by training domains $\E_{tr}$ correspond to the ones by all domains $\E$. \citet{arjovsky2020} also  discussed the relationship between the equation $\I_{tr} = \I$ and o.o.d.~generalization, briefly illustrating  that the equation $\I_{tr} = \I$ facilitates the estimation of a predictor with high o.o.d.~performance solely based on data from training domains $\E_{tr}$ \citep[Section 4.1]{arjovsky2020}. If it holds that $\I_{tr} = \I$,  we can capture an invariance $\Phi \in \I$ among all domains $\E$ only using the training domains $\E_{tr}$. \citet{arjovsky2020} pointed that, once an invariance $\Phi \in \I$ among all domains is obtained, a predictor $w^*$ that minimizes risks only on training domains $\E_{tr}$, namely $w^* \in \argmin_\W \sum_{e \in \E_{tr}} \R^e (w \circ \Phi)$,  satisfies $\R^e (w^* \circ \Phi) = \min_w \R^e (w \circ \Phi)$  for all domains $e \in \E$,  including unseen domains $\E - \E_{tr}$, under certain settings.  Developing the discussion by \cite{arjovsky2020}, Theorem \ref{theo:LSLsetting} clarifies a more rigorous relation among the equation $\I_{tr} = \I$, the o.o.d.~risk (\ref{eq:OOD-Bayes rule}), and the bi-leveled optimization problem (\ref{eq:IRM_conti}): the equation $\I = \I_{tr}$ is one of the sufficient conditions  for the bi-leveled optimization problem (\ref{eq:IRM_conti}) to minimize the o.o.d. risk (\ref{eq:OOD-Bayes rule}).

The condition $\I_{tr} = \I$ is not generally satisfied and \cite{peters2016, arjovsky2020} presented sufficient conditions  on the training domains $\E_{tr}$ for the equation $\I_{tr} = \I $ when data follow simple SEMs. \cite{peters2016} proved the equation $\I_{tr} = \I$ holds when distributions on domains follow a linear Gaussian SEM and training data are obtained by certain types of interventions \citep[Section 4.3]{peters2016}.  \citet{arjovsky2020} generalized the result by \cite{peters2016}. Assuming that data follow a linear SEM,   
which is not restricted to a Gaussian distribution and a certain type of interventions, \citet{arjovsky2020} deduced a sufficient condition for the equality $\I_{tr} = \I$ on training domains $\E_{tr}$, which is called {\em lying in the general position} \citep[Assumption 8]{arjovsky2020}. On the other hand, 
sufficient conditions for the equality $\I_{tr} = \I$ under the  setting (\ref{theo:assump1}) have not yet been revealed. Providing them would be an important area for future research.
 
\paragraph{Conditions (ii) and (iii): }  As shown in Lemma \ref{Lem:nonlin_rojas}, the conditional expectation $\int y \cdot p^I(y|x_1 )dy = \bE[Y^I =y | X_1^I = x_1] $ achieves the minimization of the o.o.d.~risk, signifying that the information embedded in $\X_1$ is important  for predicting response variables on unseen domains. Condition (ii) implies that the support of training domains $\E_{tr}$ covers $\X_1$ that contains such important information for o.o.d.~prediction. Condition (iii) implies that $\H$ is such a large feature space that a feature $\Phi: \X \rightarrow \H$  can preserve information on the $\X_1$-component of $x \in \X$ by selecting $\Phi$ appropriately. Condition  (iii) also provides a practical perspective on how to construct the feature space $\H = \bR^{d_{\H}}$: the dimension $d_{\H}$ on the feature space $\H$ should be fixed high. \st{The dimension $d_{\H}$ of the feature space is fixed by hand, and hence, Condition (iii) is expected to hold unless we fix the dimension of the feature space too small. 
}

\paragraph{Condition (iv): } Condition (iv) presents continuity of the p.d.f.~of $P_{Y^I | X_1^I}$. By Condition (iv), we also have  continuity of the conditional expectation $\int y \cdot p^I(y|x_1 )dy = \bE[Y^I =y | X_1^I = x_1] $. In our analysis, we assume that the model $\W^{\C_0}$ consists of all continuous functions, and hence, Condition (iv) ensures that the model includes the conditional expectation $\bE[Y^I|X_1^I]$, which minimizes the o.o.d.~risk (Lemma \ref{Lem:nonlin_rojas}).

%-------------------------------------------------------------------------------------------------------------------------------------------------------------------

%-------------------------------------------------------------------------------------------------------------------------------------------------------------------
 \subsection{Case II: Cross Entropy  Loss}\label{subsec:theo_in_CE}

Theorem 1 can be easily extended to  the classification case where $w \in \W$ has a probabilistic  output and evaluate risks by the cross entropy loss. Let $\Y$ be a finite set $\{1,...,m \}$ ($m \in \bN_{>0}$), and we model $w: \H \rightarrow \Y$ by $p_{\theta}: \H \rightarrow \P_{\Y}$, where  $\P_{\Y}$ denotes \kf{the set} of probabilities on $\Y$; namely
$$\P_{\Y} :=   \left\{ p \in  \bR_{+}^m \left| \sum_{i=1}^m p_i = 1 \right. \right\}.$$ Here,   $\bR_{+} := \left\{ x \in \bR \left| x \geq 0 \right. \right\}$ and $p_i$ denotes the $i$-th component of $p$. We call $p_\theta : \H \rightarrow \P_{\Y} $ continuous, that is $p_\theta \in  \W^{\C_0}$,  when correspondence $\H \ni h  \longmapsto p_\theta(h) \in \bR^{| \Y |}$ is continuous, seeing $p_\theta(h) \in \P_{\Y}$ as a vector on $\bR^{| \Y |}$. For a given $p_{\theta}: \H \rightarrow \P_{\Y}$ and $i \in \Y$, $\bigl( p_{\theta}(h) \bigr)_i$ is  often abbreviated  by $p_\theta (i | h)$.
\kff{The risk evaluated by the cross-entropy loss is then written as}
$$
\R^{e}(p_{\theta} \circ \Phi) = \int   - \log  p_{\theta} ( Y^e | \Phi(X^e)) dP_{X^e, Y^e}.
$$

We expand Theorem 1 to the above classification case:

\begin{theo}[o.o.d.~optimality of the bi-leveled optimization problem  (\ref{eq:IRM_conti}) under cross-entropy loss setting]\label{theo:CEsetting}
Domains  $\{ (X^e, Y^e)\}_{e \in \E}$ are assumed to be (\ref{theo:assump1}). Assume that, in addition to (i) $\sim$ (iii) in Theorem \ref{theo:LSLsetting}, the following condition (v) holds:
\begin{itemize}
    \item[(v)] For any $x_1^* \in \X_1$, 
    $ \# \left\{ y \in \Y \left| p^I (y | x_1^*) > 0 \right. \right\} > 1.$
\end{itemize}

 Then, we have the inclusion
\begin{equation}\label{eq:IRMoptimal_CE}
 \argmin \nolimits_{\Phi \in \I_{tr}^{\C_0}, \\
p_\theta \in \W^{\C_0}}  \sum_{e \in \E_{tr}} \R^{e}( p_\theta \circ \Phi ) \subset \argmin_{p_\theta \in \F} \st{ \R^{o.o.d.}(f)},
\end{equation}
 where $\F$ is the set of all measurable functions \st{$f: \X \rightarrow \P_{\Y}$.}\footnote{\st{The same as the definition of {\em continuous}, we call \st{$f \in  \F$} measurable when correspondence $\X \ni x  \longmapsto \st{f(x)} \in \bR^{| \Y |}$ is measurable, seeing $\st{f(x)} \in P_{\Y}$ as a vector on $\bR^{| \Y |}$. }} 
\end{theo}

Condition (v) indicates that domains $ \{ (X^e, Y^e) \}_{e \in \E}$ have high uncertainty in labels $y \in \Y$ given $x_1 \in \X_1$. The condition is expected to be feasible  when classes $\Y$ are subdivided and difficult to be uniquely determined from $x_1 \in \X_1$.

\subsection{Novelty and Significance of Theorems 1 and 2 }\label{subsec:nov_and_sig}

Theorems 1 and 2 and their proofs have the following \st{four} novel and  significant points:

\paragraph{Setting of Domains} The first point is the setting of domains. The setting by \cite{rojas-carulla2018a}, which is used throughout our analysis, does not impose any specific SEM structures, linearity, and  Gaussianity on domains while existing works on theoretical analysis of IRM assumed that data follow simple SEMs. Theorems 1 and 2 indicate that, under such a general setting, IRM presents the minimum of the o.o.d.~risk. This implies that our results provide a solid foundation to use IRM for a broad range of o.o.d.~generalization problem.

\paragraph{\st{Assumption on the Underlying Distribution $P_{Y^I|X_1^I}$}} \st{As well as the assumption on domains, prior theoretical results of IRM assume that the underlying true distribution $P_{Y^I |X_1^I}$ is represented by  a simple SEM \citep{arjovsky2020, rosenfeld2021, kamath2021}. On the other hand, Condition (iv) only imposes $P_{Y^I|X_1^I}$ on the continuity; hence, Condition (iv) is a significantly mild condition in comparison with the assumptions on $P_{Y^I |X_1^I}$  by prior works.}

\paragraph{Characterization of Invariance}Second,  a theoretical characterization of invariances $\Phi^* \in \I^{\C_0}$ is given in Lemmas \ref{Lem:invaiance=compiose} and \ref{Lem:invaiance=compiose_CE}: it is proved that $\Phi^* \in \I^{\C_0}$ can be represented as $\Phi^* = \Psi^* \circ \Phi^{\X_1}$ for some continuous map $\Psi^*$. Any theoretical characterizations have not yet been presented, and hence, the results in  Lemmas \ref{Lem:invaiance=compiose} and \ref{Lem:invaiance=compiose_CE} are novel.  To present  the non-trivial characterization, we develop a novel theoretical technique  based on the proof by contradiction. Lemmas \ref{Lem:invaiance=compiose} and \ref{Lem:invaiance=compiose_CE} play an important role in our desirable assertion (\ref{eq:IRMoptimal}), and hence, the derivation of these lemmas is a significant technical contribution of our analysis. 

\paragraph{Range of Invariance}
The fourth point is a range of invariances $\Phi$: we assume that $\Phi$ run all continuous functions, while most of the  existing works on theoretical analysis of IRM assume that $\Phi$ run more simplified functions, such as linear functions \citep{rosenfeld2021} or variable selections \citep{toyota2022}.
It is common to construct a learning model of invariances with deep neural networks in the context of IRM, and hence, the variable selection and linear function settings by \cite{toyota2022, rosenfeld2021} are significantly simplified  to analyze IRM. On the other hand, our large class of continuous functions is relatively realistic compared to existing ones, since it is widely recognized that neural networks of sufficient size can represent a wide range of functions \citep{cybenko1989, hornik1989, barron1993, mhaskar1996, sonoda2017}.

%-------------------------------------------------------------------------------------------------------------------------------------------------------------------

\section{Previous Works}\label{sec:related}

As explained in Section \ref{sec:intro}, \citet{rosenfeld2021, kamath2021} derived the theoretical results concerning the minimum of the bi-leveled optimization problem (\ref{eq:IRM}) and its connection to the o.o.d.~risk (\ref{eq:OOD-Bayes rule}). \citet{rosenfeld2021} proved that a predictor obtained by minimizing (\ref{eq:IRM}) predicts $Y^e \in \Y$ relying only on a feature of $X^e \in \X$ whose distribution does not depend on domains \citep[Section 5]{rosenfeld2021}. However,  they did not provide any connections between the minimum of (\ref{eq:IRM}) and the o.o.d~risk. Moreover, they assume that data follow a linear Gaussian SEM, and that invariances $\Phi$ in the bi-leveled optimization problem run linear functions for simplicity.  Unlike their analysis, this paper derives the direct relations between (\ref{eq:IRM}) and the o.o.d.~risk (\ref{eq:OOD-Bayes rule}). Additionally, we assume that data follow the distributions proposed by \citet{rojas-carulla2018a} that do not rely on any specific SEM structures and that invariances run all continuous functions including neural networks. \citet{kamath2021} provided an example of distributions on which a minimum of (\ref{eq:IRM}) does not minimize the o.o.d~risk.  However,   the distributions are particular SEMs constructed to derive the case where (\ref{eq:IRMoptimal}) is violated, and analysis in more general settings is required \citep[Section 4]{kamath2021}. In construct, the distributions by \citet{rojas-carulla2018a} used in this paper do not rely on any specific SEM structures, and  
they are used to analyze estimation methods related to  invariances \citep{rojas-carulla2018a, toyota2022}.

\citet{arjovsky2020, koyama2021, rojas-carulla2018a} discussed theoretical relations between invariances and the o.o.d.~risk (\ref{eq:OOD-Bayes rule}). As explained in the last section, \citet{arjovsky2020} intuitively explained that the condition  $\I_{tr} = \I$ facilitates an estimation of a predictor which can predict $Y^e$ on unseen domains only by data from training domains $\E_{tr}$ \citep[Section 4.1]{arjovsky2020}. They also derived sufficient conditions on training domains for  the equation $\I_{tr} = \I$, assuming that data follow a simple linear SEM \citep[Theorem 9]{arjovsky2020}. \citet{koyama2021, rojas-carulla2018a} presented sufficient conditions for an invariance $\Phi$ to achieve the minimum of (\ref{eq:OOD-Bayes rule}). \st{\citet{koyama2021} proved that the invariance  that maximizes the mutual information with labels also maximizes the o.o.d.~risk. \citet{rojas-carulla2018a} proved that, under the domain setting (\ref{theo:assump1}), the conditional expectation $\bE[Y^e |\Phi^{\X_1} (X^e) = x_1]$ also minimizes the o.o.d.~risk, even when $\bE[Y^e |\Phi^{\X_1} (X^e)]$ is nonlinear.} However, all  the results by \citet{koyama2021, rojas-carulla2018a, arjovsky2020}  did not deal with any theoretical connections between invariances obtained by minimizing the bi-leveled optimization problem (\ref{eq:IRM}) and the o.o.d.~risk (\ref{eq:OOD-Bayes rule}). \st{It does not follow obviously that the minimum of (\ref{eq:IRM}) satisfies these sufficient conditions by \citet{koyama2021, rojas-carulla2018a}, and hence our main theorems can not be deduced as a trivial corollary of the results by \citet{koyama2021, rojas-carulla2018a}. To discuss the non-trivial relation between the bi-leveled optimization problem (\ref{eq:IRM}) and the o.o.d.~risk (\ref{eq:OOD-Bayes rule}), we establish a novel characterization of invariances (Lemmas \ref{Lem:invaiance=compiose} and \ref{Lem:invaiance=compiose_CE}), and derive the main theorems based on it. }

To reduce the annotation cost required for the original IRM approach, \cite{toyota2022} introduced a new bi-level optimization problem similar to (\ref{eq:IRM}). They considered a situation in which the training data for target classification are provided in only one domain, while the task of a higher label hierarchy, which requires lower annotation cost, has data from multiple domains. Under the availability of data, they deduced a bi-level optimization problem, in which  invariances were given by additional data in a higher label hierarchy. For further details, we refer the reader to the original paper \cite{toyota2022}. Their study provided a detailed theoretical analysis concerning their method and its connection to the o.o.d.~risk; however, they did not analyze relationships  between their bi-leveled optimization problem and the o.o.d.~risk, which is the focus of this paper. Instead, they investigated relationships between an optimization method for their bi-level optimization problem and the o.o.d.~risk. Moreover, they assume that invariances $\Phi$ run all variable selections for simplicity of theoretical analysis. On the other hand, this paper derives the direct relations between the minimum of the bi-leveled optimization problem and the o.o.d.~risk (\ref{eq:OOD-Bayes rule}). Moreover, we consider the more realistic setting for the analysis of IRM where invariances $\Phi$ run all continuous functions.

%-------------------------------------------------------------------------------------------------------------------------------------------------------------------
\section{Proofs}\label{sec:proof}
%-------------------------------------------------------------------------------------------------------------------------------------------------------------------
In this section, we prove Theorem 1 and 2. Through the section, for $X^e \in \X$ and $x \in \X$, its  $\X_i$-components ($i=1, 2$) are denoted by $X^e_i$ and $x_i$ respectively.

\subsection{\textcolor{black}{Proof Sketch of Main Theorems}}
\st{
Before giving rigorous proof, we briefly describe the rough proof sketch of the main theorems. The following two lemmas (A) and (B) play an important role in our proof:
\begin{itemize}
\item[(A)] The conditional expectation $\bE[Y^e | X_1^e] = \bE[Y^I | X_1^I]$ and conditional probability $P_{Y^e | X_1^e} = P_{Y^I |X_1^I}$ minimize the o.o.d.~risk under the least-square and cross-entropy losses respectively (Lemmas \ref{Lem:nonlin_rojas} and \ref{Lem:nonlin_rojas_CE}).
\item[(B)] Any invariance among all domains can be represented by the composition of the projection onto $\X_1$; that is, $ \Phi \in  \I^{\C_0}$  can be represented as $$  \Phi  = \Psi \circ \Phi^{\X_1} $$
for some continuous map $\Psi$ (Lemmas \ref{Lem:invaiance=compiose} and \ref{Lem:invaiance=compiose_CE}).
\end{itemize}
The  two lemmas intuitively conclude the proof of the main theorem as follows. Firstly,  since $\I^{\C_0}_{tr} = \I^{\C_0}$ holds (Condition (i)), observe that a predictor in (\ref{eq:IRM}) runs composition maps $w \circ \Phi$ with $w \in \W^{\C_0}$ and $\Phi \in \I^{\C_0}$. Moreover, since the above second lemma (B) ensures that $\Phi \in \I^{\C_0}$ can be represented by the composition of the projection onto $\X_1$, we can see that  a predictor in (\ref{eq:IRM}) runs $w \circ \Phi^{\X_1}$ for some function class $w \in \W^*$, and hence, the bi-leveled optimization problem is expressed as 
\begin{equation}
 \min \nolimits_{ \\
w \in \W^*}  \sum_{e \in \E_{tr}} \R^{e}(w \circ \Phi^{\X_1} ) .
\end{equation}
It is well-known that, assuming that $w$ runs all measurable functions, 
\begin{equation*}
    \hat{w} \in \argmin_w \R^{e} (w \circ \Phi^{\X_1}) \Longleftrightarrow \hat{w}(x_1) = \bE[Y^e | X_1^e =x_1] = \bE[Y^I | X_1^I=x_1] ~~P_{X^e_1}-\text{almost everywhere}. %\footnote{While the inverse implication also holds, it is not utilized in our proof.}
\end{equation*}
or 
\begin{equation*}
    \hat{w} \in \argmin_w \R^{e} (w \circ \Phi^{\X_1}) \Longleftrightarrow \hat{w}(x_1) = P_{Y^e | X_1^e=x_1} = P_{Y^I |X_1^I=x_1} ~~P_{X^e_1}-\text{almost everywhere}. %\footnote{While the inverse implication also holds, it is not utilized in our proof.}
\end{equation*}
hold for any $e \in \E$ under the least-square and cross-entropy losses respectively \citep[Example 2.6]{andreas2008}. Hence, ignoring the capability of $\W^*$ and the discussion of {\em almost everywhere}, we can see that
\begin{equation}\label{rough: IRM_MSE}
\bE[Y^I | X_1^I=x_1] \approx \argmin \nolimits_{ \\
w \in \W^*}  \sum_{e \in \E_{tr}} \R^{e}(w \circ \Phi^{\X_1} ) 
\end{equation}
or
\begin{equation}\label{rough: IRM_CE}
 P_{Y^I |X_1^I=x_1}  \approx \argmin \nolimits_{ \\
w \in \W^*}  \sum_{e \in \E_{tr}} \R^{e}(w \circ \Phi^{\X_1} ) 
\end{equation}
hold. %\footnote{\st{We can hese $\approx$ strictly correspond to $=$ by Conditions (ii), (iv) and the fact that $w \in \W^{\C_0}$ runs {\em all} continuous maps. In detail, please see the rigorous proof in the next subsections.} }.
Combining eq.s (\ref{rough: IRM_MSE}), (\ref{rough: IRM_CE}) and the first lemma (A), it concludes the main theorems intuitively. In the following section, we give the rigorous justification of the above rough proof sketch.
}

%-------------------------------------------------------------------------------------------------------------------------------------------------------------------
\subsection{Proof of Theorem \ref{theo:LSLsetting}} 
%-------------------------------------------------------------------------------------------------------------------------------------------------------------------
To prove the main theorem, we prepare two lemmas. 
\begin{lemm}\label{Lem:nonlin_rojas}
Let $w^I : \X_1 \rightarrow \Y$ be the conditional expectation obtained by $p^I (y |x_1)$; namely, 
$$w^I (x_1) =  \bE [Y^I | X_1^I = x_1]:= \int y \cdot  p^I (y |x_1) dy. $$
Then, 
$$w^I \circ \Phi^{\X_1} \in \argmin_{f: \X \rightarrow \Y}  \R^{o.o.d.}(f).$$

\end{lemm}

\begin{lemm}\label{Lem:invaiance=compiose}
Any $\Phi \in \I_{tr}^{C^0}$ is represented as 
$$  \Phi  = \Psi \circ \Phi^{\X_1} $$
for some continuous map $\Psi : \X_1 \rightarrow \H$.

\end{lemm}

\paragraph{\bf{Proof of Lemma \ref{Lem:nonlin_rojas}}}\footnote{The proof is essentially the same as the one for Theorem 1 in \cite{rojas-carulla2018a} and Theorem 6 in \cite{toyota2022}.}
It suffices to prove the following statement:
\begin{center}
    For any $f \in \F$ and $(X^{a}, Y^{a}) \in \{(X^e, Y^e) \}_{e \in \mathcal{E}}$, there exists $(X^{b}, Y^{b}) \in  \{(X^e, Y^e) \}_{e \in \mathcal{E}}$ such that 
\end{center}
\begin{equation}\label{eq1::IRM=o.o.d.}
\int \| w^I \circ \Phi^{\X_1} (x) -y \|^2 dP_{X^{a}, Y^{a}} (x, y) \leq \int \|f(x)-y \|^2 dP_{X^{b}, Y^{b}} (x, y). 
\end{equation}
Take arbitrary $f \in \F$ and $(X^{a},  Y^{a}) \in \{(X^e, Y^e) \}_{e \in \mathcal{E}}$.
Define $(X^{b}, Y^{b}) \in \{(X^e, Y^e)\}_{e\in \E}$ such that its distribution is the direct product $P_{X_1^{a}, Y^{a}} \otimes P_{X_2}$, where $P_{X_1^{a}, Y^{a}}$ is the marginal distribution of $P_{X^{a}, Y^{a}}$ on $\X_1 \times \Y$ and $P_{X_2}$ is an arbitrary distribution on $\X_2$.  

Then, the right-hand side of the inequality (\ref{eq1::IRM=o.o.d.}) is given by
\begin{align*}
    \int \|f(x)-y \|^2 dP_{X^{b}, Y^{b}} (x, y) &= \int \|f(x)-y \|^2 d(P_{X_1^{a}, Y^{a}} \otimes P_{X_2}) (x, y) \\
    & = \int  P_{X_2}(x_2) \int  \|f(x_1,x_2)-y \|^2 dP_{X_1^{a}, Y^{a}} (x_1, y).
\end{align*}
Clearly,  for any $x_2^* \in \X_2$, the inequality 
\begin{align*}
\int  \|f(x_1,x_2^*)-y \|^2 dP_{X_1^{a}, Y^{a}} (x_1, y) &\geq \int  \|\mathbb{E}[Y^{e_1} | X^{a}_1 = x_1]-y \|^2 dP_{X_1^{a}, Y^{a}} (x_1, y) 
\end{align*}
holds, because the  minimum of a risk on the least square loss is attained at the conditional expectation $\mathbb{E}[Y^{a} | X^{a}_1]$. Hence, we obtain 
\begin{align*}
    \int \|f(x)-y \|^2 dP_{X^{b}, Y^{b}} (x, y) &= \int  P_{X_2}(x_2) \int  \|f(x_1,x_2)-y \|^2 dP_{X_1^{a}, Y^{a}} (x_1, y) \\
    & \geq \int  P_{X_2}(x_2) \int  \|\mathbb{E}[Y^{e_1}| X^{a}_1 = x_1]-y \|^2 dP_{X_1^{a}, Y^{a}} (x_1, y)  \\
    & = \int  \|\mathbb{E}[Y^{a} | X^{a}_1 = x_1]-y \|^2 dP_{X_1^{a}, Y^{a}} (x_1, y)  \\
    & = \int  P_{X^{a}_2| X^{a}_1, Y^{a}}(x_2)  \int  \|\mathbb{E}[Y^{a} | X^{a}_1 = x_1]-y \|^2 dP_{X_1^{a}, Y^{a}} (x_1, y)  \\
    %& =  \int  \|\mathbb{E}[Y^{a} | \Phi^{\X_1}(X^{a}) = x_1]-y \|^2  d(P_{X_1^{a}, Y^{a}} \otimes P_{X^{a}_2| X^{a}_1, Y^{a}})(x_1, x_2,  y) \\
    %& =  \int \|\mathbb{E}[Y^{a} | \Phi^{\X_1}(X^{a}) = x_1]-y \|^2  dP_{X^{a}, Y^{a}} (x, y) \\
    & =  \int \|\mathbb{E}[Y^{a} | X^{a}_1 =  \Phi^{\X_1}(x)]-y \|^2  dP_{X^{a}, Y^{a}} (x, y) \\
     & =  \int  \| w^I \circ \Phi^{\X_1}(x)-y \|^2 dP_{X^{a}, Y^{a}} (x, y),
\end{align*}
which concludes the proof. Here, the last equality is derived from the fact that the conditional expectation $\mathbb{E}[Y^{e} | X^{e}_1 =  \Phi^{\X_1}(x)]$ does not depend on $e \in \E$ and corresponds to $w^I \circ \Phi^{\X_1}$.
\qed

\paragraph{\bf{Proof sketch of Lemma \ref{Lem:invaiance=compiose}}}
Before providing a complete proof, we show a proof sketch of Lemma \ref{Lem:invaiance=compiose} to make the flow of our proof easier to understand. First, we prove that $\Phi \in \I_{tr}^{C^0}$ can be represented as 
\begin{equation}\label{min_compose}
\Phi = \Psi \circ \Phi^{\X_1}
\end{equation}
by some map $\Psi : \X_1 \rightarrow \H$, which is not restricted to a continuous map. Take arbitrary $\Phi \in \I_{tr}^{C^0}$. 
Then, since $\Phi \in \I^{C^0} =\I_{tr}^{C^0}$ (Condition (i)),  for any $(X^a, Y^a), (X^b, Y^b) \in \{ (X^e, Y^e) \}_{e \in \E}$,
\begin{equation*}
P_{Y^a  | \Phi (X^a)} = P_{Y^b  | \Phi (X^b) }, 
\end{equation*}
and therefore, we have
\begin{equation}\label{phi_inv}
%P_{Y^a \in N | \Phi(X^a) =\Phi(x)} = P_{Y^b \in N | \Phi(X^b) =\Phi(x)}
P_{Y^a | \Phi(X^a)}(N |\Phi(x) ) = P_{Y^b | \Phi(X^b) } (N |\Phi(x) )
\end{equation}
for any set $N \subset \Y$ and $\forall x \in \X$. We prove the statement (\ref{min_compose}) by contradiction.  Assume that there exist no maps $\Psi$ that satisfy (\ref{min_compose}). Then, there exist $x_1^* \in \X_1$, $x_2^*, x_2^{**} \in \X_2$ such that 
$$  \Phi (x_1^*, x_2^*) \neq \Phi(x_1^*, x_2^{**}).\footnote{\textcolor{black}{If $\Phi (x_1^*, x_2^*) = \Phi(x_1^*, x_2^{**})$ for any $x_1^* \in \X_1$, $x_2^*, x_2^{**} \in \X_2$, $\Phi$ depend only on the first component $\X_1$; hence, we can see that $\Phi \in \I_{tr}^{C^0}$ can be represented as 
$
\Phi = \Psi \circ \Phi^{\X_1}
$
by some map $\Psi$, which contradicts to the assumption. }}$$ By utilizing $x_1^* \in \X_1$, $x_2^*, x_2^{**} \in \X_2$, we can construct $(X^a, Y^a), (X^b, Y^b) \in \{ (X^e, Y^e) \}_{e \in \E}$ and $N \subset \Y$ which satisfy 
\begin{equation*}
P_{Y^a  | \Phi  (X^a) } (N |\Phi (x_1^*, x_2^*)  ) \neq P_{Y^b  | \Phi(X^b)} (N |\Phi (x_1^*, x_2^*)  ).
\end{equation*}
This contradicts the assumption (\ref{phi_inv}), and we can conclude $\Phi \in \I_{tr}^{C^0}$ can be represented as (\ref{min_compose}). The continuity of $\Psi$ is easily derived from the continuity of $\Phi$, and we can conclude the proof.

\paragraph{\bf{Proof of Lemma \ref{Lem:invaiance=compiose}}}
First, we prove that $\Phi \in \I_{tr}^{C^0}$ can be represented as 
\begin{equation}\label{min_compose2}
\Phi = \Psi \circ \Phi^{\X_1}
\end{equation}
by some map $\Psi  : \X_1 \rightarrow \H$, which is not restricted to a continuous map. We prove this statement by contradiction. Take $\Phi \in \I_{tr}^{C^0}$ and assume that there exist no  maps $\Psi$ which satisfy  (\ref{min_compose2}). Then, there exist $x_1^* \in \X_1$, $x_2^*, x_2^{**} \in \X_2$ such that 
\begin{equation}\label{eq:assumption_contrudiction}
\Phi (x_1^*, x_2^*) \neq \Phi(x_1^*, x_2^{**}). 
\end{equation} 
Fix $y^* \in \Y$ with $p^I(y^*|x_1^*) > 0$ and  take an open neighborhood $N_{y^*}  \subset \Y$ centered at $y^*$ which satisfies 
$$ 0 < \int_{N_{y^*}} p^I(y|x^*_1) dy < 1. $$ Here, the existence of  $N_{y^*}$ is derived from the continuity of $p^I(\cdot | x_1^*)$ (Condition (iv)).

Define two maps $g^i: \Y \rightarrow \X_2$ ($i=1,2$) by 
%\vspace{5mm}
 \begin{center}
  $ g^1( y) = \left\{
\begin{array}{ll}
 x_2^* & (y \in N_{y^*})\\
x_2^{**} & (\text{ else })
\end{array}
\right. $ \textcolor{white}{aaaaaa} 
$g^2( y) = \left\{
\begin{array}{ll}
  x_2^{**} & (y \in N_{y^*})\\
x_2^{*} & (\text{ else }).
\end{array}
\right. $
 \end{center}
Take two distributions $(X^{a}, Y^{a} ), (X^{b}, Y^{b} ) \in \{ (X^e, Y^e) \}_{e \in \E}$   such that their distributions $P_{X^{a}, Y^{a}}$ and   $P_{X^{b}, Y^{b}}$ coincide with 
 \begin{align*}
P_{X^{a}, Y^{a}} = P_{X_2^{a} | Y^{a}} \otimes  P_{Y^I | X_1^I }
\otimes 
P_{X_1}, ~~~~~
P_{X^{b}, Y^{b}} =P_{X_2^{b} | Y^{b}}  \otimes  P_{Y^I | X_1^I } 
\otimes 
P_{X_1}.
\end{align*}
\begin{figure}[t]
\centering 
\includegraphics[width = 0.5\textwidth]{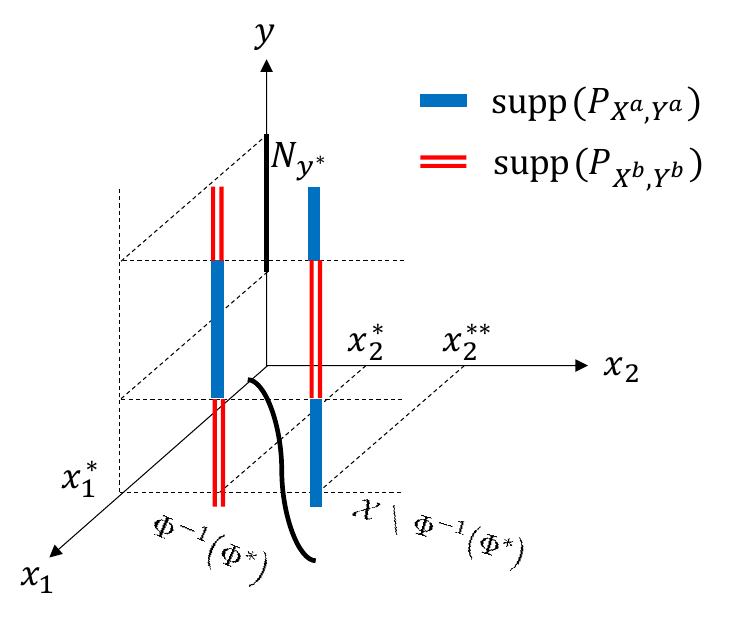}
 %¥vspace{-4mm}
\caption{Supports of probability distributions $P_{X^a, Y^a}$ and $P_{X^b, Y^b}$. The figure implies that $P_{X^a, Y^a}(N_{y^*} \times \Phi^{-1} (\Phi^*)) \neq 0$ and $P_{X^b, Y^b}(N_{y^*} \times \Phi^{-1} (\Phi^*)) = 0$ ($\because (x_1^*, x_2^{**}) \notin \Phi^{-1} (\Phi^*)$ (\ref{eq:assumption_contrudiction})), and that $P_{X^a, Y^a}(\Phi^{-1} (\Phi^*)) \neq 0$ and $P_{X^b, Y^b}(\Phi^{-1} (\Phi^*)) \neq 0$. These Eqs.~lead us  $P_{Y^{a} |\Phi(X^{a})} (N_{y^*} | \Phi^*) \neq 0 = P_{Y^{b} |\Phi(X^{b})} (N_{y^*} | \Phi^*)$. }\label{fig:vizualized_probs}
%¥vspace{-2mm}
\end{figure}
Here, 
\begin{itemize}
\item  $P_{X_1 }$ is a  distribution on $\X_1$  where its p.d.f. coincides with  a delta function $ \delta_{x_1^*} (x_1)$ on $x_1^*$,
\item the conditional p.d.f.s of $P_{X_2^{a} | Y^{a}}(\cdot | y)$ and $P_{X_2^{b} | Y^{b}}(\cdot | y)$    coincide with $\delta_{g^1(y)} (x_2)$ and  $ \delta_{g^{2}(y)} (x_2) $ respectively.
\end{itemize} 
The supports of $P_{X^a, Y^a}$ and $P_{X^b, Y^b}$ are visualized in Fig.~\ref{fig:vizualized_probs}.
As $\Phi \in \I^{\C_0} = \I^{\C_0}_{tr}$ (Condition (i)) and $(X^{a}, Y^{a} ), (X^{b}, Y^{b} ) \in \{ (X^e, Y^e) \}_{e \in \E}$,
\begin{equation}\label{eq:invariance}
P_{Y^{a} |\Phi(X^{a})} (N_{y^*} | \Phi^*)=P_{Y^{b} |\Phi(X^{b})} (N_{y^*} | \Phi^*), 
\end{equation}
where $ \Phi^* := \Phi(x_1^*, x_2^*)$.  Let us compute $P_{Y^{a} |\Phi(X^{a})} (N_{y^*} | \Phi^*)$ and $P_{Y^{b} |\Phi(X^{b})} (N_{y^*} | \Phi^*)$ to derive $P_{Y^{a} |\Phi(X^{a})} (N_{y^*} | \Phi^*) \neq P_{Y^{b} |\Phi(X^{b})} (N_{y^*} | \Phi^*)$, which contradicts
to the equality (\ref{eq:invariance})\footnote{Fig.~\ref{fig:vizualized_probs} illustrates the intuitive reason why $P_{Y^{a} |\Phi(X^{a})} (N_{y^*} | \Phi^*) \neq P_{Y^{b} |\Phi(X^{b})} (N_{y^*} | \Phi^*)$ is derived, and hence, will help us understand the following rigorous proof.}.  We evaluate $$P_{Y^{a} |\Phi(X^{a})} (N_{y^*} | \Phi^*) = \frac{ P_{\Phi(X^{a}), Y^{a}} ( \{\Phi^*\} \times N_{y^*} )}{P_{\Phi(X^{a})} ( \{\Phi^*\}  )}  = \frac{ P_{X^{a}, Y^{a}} ( \Phi^{-1} (\Phi^*) \times N_{y^*} )}{P_{X^{a}} ( \Phi^{-1} (\Phi^*)) } $$
by computing its numerator and denominator separately.  First, the numerator is evaluated as 
\begin{align*}
 P_{X^{a}, Y^{a}} ( \Phi^{-1} (\Phi^*) \times N_{y^*} )&= \int_{ \Phi^{-1} (\Phi^*) \times N_{y^*}}   \delta_{g^1(y)} (x_2) \cdot p^I(y |x_1) \cdot   \delta_{x_1^*} (x_1) dxdy \\
 &= \int_{N_{y^*}} dy\int_{\Phi^{-1} (\Phi^*)}\delta_{g^1(y)} (x_2)  \cdot p^I(y |x_1) \cdot   \delta_{x_1^*} (x_1) dx.
 \end{align*}
Noting that $g^{1}(y) = x_2^{*}$ for $\forall y \in N_{y^*}$ and $\delta_{x_1^*} (x_1) \times \delta_{x_2^*} (x_2) = \delta_{(x_1^*, x_2^*)} (x_1, x_2)$, we obtain 
\begin{align*}
 \int_{N_{y^*}} dy\int_{\Phi^{-1} (\Phi^*)}\delta_{g^1(y)} (x_2)  \cdot p^I(y |x_1) \cdot   \delta_{x_1^*} (x_1) dx 
 &=   \int_{N_{y^*}} dy\int_{\Phi^{-1} (\Phi^*)}  \delta_{x_2^*} (x_2) \cdot p^I(y |x_1) \cdot \delta_{x_1^*} (x_1) dx   \\
 &=  \int_{N_{y^*}} dy\int_{\Phi^{-1} (\Phi^*)}  \delta_{(x_1^*, x_2^*)} (x_1, x_2) \cdot p^I(y |x_1)  dx.
 \end{align*}
Since  $(x_1^*, x_2^{*})  \in \Phi^{-1} (\Phi^*)$,  we have
 $$ \int_{N_{y^*}} dy\int_{\Phi^{-1} (\Phi^*)}  \delta_{(x_1^*, x_2^*)} (x_1, x_2)  \cdot p^I(y |x_1)  dx = \int_{N_{y^*}} p^I(y |x_1^*)  dy,  $$
 which leads us to the equality
 $$ P_{X^{a}, Y^{a}} ( \Phi^{-1} (\Phi^*) \times N_{y^*} )  =  \int_{N_{y^*}} p^I(y |x_1^*)  dy. $$

Next, let us evaluate the denominator $P_{X^{a}} ( \Phi^{-1} (\Phi^*))$. 
\begin{align*}
 P_{X^{a}} ( \Phi^{-1} (\Phi^*)) &= P_{X^{a}, Y^a} ( \Phi^{-1} (\Phi^*) \times \Y) =  \int_{ \Phi^{-1} (\Phi^*)  \times \Y }  \delta_{g^1(y)} (x_2)  \cdot  p^I(y |x_1) \cdot   \delta_{x_1^*} (x_1) dxdy   \\
 &=\int_{\Y} dy\int_{\Phi^{-1} (\Phi^*)}\delta_{g^1(y)} (x_2) \cdot  p^I(y |x_1) \cdot   \delta_{x_1^*} (x_1) dx  \\
 &=  \int_{N_{y^*}} dy\int_{\Phi^{-1} (\Phi^*)}  \delta_{g^1(y)} (x_2) \cdot p^I(y |x_1)  \cdot \delta_{x_1^*} (x_1) dx   \\
 &~~~~~~~~~~~~~~~~~~~~~~~~~~~~~~+   \int_{\Y - N_{y^*}} dy\int_{\Phi^{-1} (\Phi^*)}\delta_{g^{1}(y)} \cdot p^I(y |x_1)\cdot  \delta_{x_1^*} (x_1) dx  \\
  &=  \int_{N_{y^*}} dy\int_{\Phi^{-1} (\Phi^*)}  \delta_{x_2^*} (x_2) \cdot p^I(y |x_1)\cdot  \delta_{x_1^*} (x_1) dx    \\
 &~~~~~~~~~~~~~~~~~~~~~~~~~~~~~~+  \int_{\Y - N_{y^*}} dy\int_{\Phi^{-1} (\Phi^*)}\delta_{x_2^{**}} (x_2) \cdot p^I(y |x_1) \cdot \delta_{x_1^*} (x_1) dx \\
 &=  \int_{N_{y^*}} dy\int_{\Phi^{-1} (\Phi^*)}  \delta_{(x_1^*, x_2^*)} (x_1, x_2) \cdot p^I(y |x_1) dx    \\
 &~~~~~~~~~~~~~~~~~~~~~~~~~~~~~+  \int_{\Y - N_{y^*}} dy\int_{\Phi^{-1} (\Phi^*)}\delta_{(x_1^*, x_2^{**})}(x_1, x_2) \cdot p^I(y |x_1)  dx. 
 \end{align*}
 Here, the fourth equality is derived from the facts that $g^{1}(y) = x_2^{*}$ for $\forall y \in N_{y^*}$ and $g^{1}(y) = x_2^{**}$ for $\forall y \in \Y - N_{y^*}$.
Noting that $(x_1^*, x_2^{*})  \in \Phi^{-1} (\Phi^*)$ and $(x_1^*, x_2^{**})  \notin \Phi^{-1} (\Phi^*)$, we obtain 
 \begin{align*}
 P_{X^{a}} ( \Phi^{-1} (\Phi^*)) &=  \int_{N_{y^*}} dy\int_{\Phi^{-1} (\Phi^*)}  \delta_{(x_1^*, x_2^*)} (x_1, x_2) \cdot p^I(y |x_1) dx    \\
 &~~~~~~~~~~~~~~~~+  \int_{\Y - N_{y^*}} dy\int_{\Phi^{-1} (\Phi^*)}\delta_{(x_1^*, x_2^{**})}(x_1, x_2) \cdot p^I(y |x_1)  dx \\
 &=  \int_{N_{y^*}} dy \cdot p^I(y |x_1^*)     
 +  \int_{\Y - N_{y^*}} dy \cdot 0 \\
 &=  \int_{N_{y^*}}  p^I(y |x_1^*)  dy .
 \end{align*}

Hence, we obtain  
 \begin{align*}
  P_{Y^{a} |\Phi(X^{a})} (N_{y^*} | \Phi^*) &= \frac{ P_{X^{a}, Y^{a}} ( \Phi^{-1} (\Phi^*) \times N_{y^*} ) }{P_{X^{a}} ( \Phi^{-1} (\Phi^*)) }  \\ \\
 &= \frac{ \int_{N_{y^*}}  p^I(y |x_1^*)  dy}{ \int_{N_{y^*}}  p^I(y |x_1^*)  dy} = 1.
 \end{align*}
 Next, let us evaluate
 $$P_{Y^{b} |\Phi(X^{b})} (N_{y^*} | \Phi^*) =  \frac{ P_{\Phi(X^{b}), Y^{b}} ( \{\Phi^*\} \times N_{y^*} )}{P_{\Phi(X^{b})} ( \{\Phi^*\} ) }  =  \frac{P_{X^{b}, Y^{b}} ( \Phi^{-1} (\Phi^*) \times N_{y^*} ) }{P_{X^{b}} ( \Phi^{-1} (\Phi^*)) } .$$
 The numerator is evaluated as 
\begin{align*}
P_{X^{b}, Y^{b}} ( \Phi^{-1} (\Phi^*) \times N_{y^*} ) &= \int_{    \Phi^{-1} (\Phi^*) \times N_{y^*}}   \delta_{g^2(y)} (x_2) \cdot p^I(y |x_1) \cdot   \delta_{x_1^*} (x_1) dxdy \\
 &= \int_{N_{y^*}} dy\int_{\Phi^{-1} (\Phi^*)}\delta_{g^2(y)} (x_2)  \cdot p^I(y |x_1) \cdot   \delta_{x_1^*} (x_1) dx .
 \end{align*}
Noting that $g^{2}(y) = x_2^{**}$ for $\forall y \in N_{y^*}$, we obtain 
\begin{align*}
 \int_{N_{y^*}} dy\int_{\Phi^{-1} (\Phi^*)}\delta_{g^2(y)} (x_2)  \cdot p^I(y |x_1) \cdot   \delta_{x_1^*} (x_1) dx 
 &=   \int_{N_{y^*}} dy\int_{\Phi^{-1} (\Phi^*)}  \delta_{x_2^{**}} (x_2) \cdot p^I(y |x_1) \cdot \delta_{x_1^*} (x_1) dx   \\
 &=  \int_{N_{y^*}} dy\int_{\Phi^{-1} (\Phi^*)}  \delta_{(x_1^*, x_2^{**})} (x_1, x_2) \cdot p^I(y |x_1)  dx  \\
 &=  \int_{N_{y^*}} dy\cdot 0 =0.
 \end{align*}
Here, the third equality is derived from $(x_1^*, x_2^{**}) \notin \Phi^{-1} (\Phi^*) $. Next,  the denominator $P_{X^{b}} ( \Phi^{-1} (\Phi^*)) $ is evaluated as
\begin{align*}
 P_{X^{b}} ( \Phi^{-1} (\Phi^*)) &= P_{X^{b}, Y^b} ( \Phi^{-1} (\Phi^*) \times \Y)  =  \int_{ \Phi^{-1} (\Phi^*)  \times \Y }  \delta_{g^2(y)} (x_2)  \cdot  p^I(y |x_1) \cdot   \delta_{x_1^*} (x_1) dxdy   \\
 &=\int_{\Y} dy\int_{\Phi^{-1} (\Phi^*)}\delta_{g^2(y)} (x_2) \cdot  p^I(y |x_1) \cdot   \delta_{x_1^*} (x_1) dx  \\
 &=  \int_{N_{y^*}} dy\int_{\Phi^{-1} (\Phi^*)}  \delta_{g^2(y)} (x_2) \cdot p^I(y |x_1)  \cdot \delta_{x_1^*} (x_1) dx  + \\
 &~~~~~~~~~~~~~~~~~~~~~~~~~~~   \int_{\Y - N_{y^*}} dy\int_{\Phi^{-1} (\Phi^*)}\delta_{g^{2}(y)} \cdot p^I(y |x_1)\cdot  \delta_{x_1^*} (x_1) dx  \\
  &=  \int_{N_{y^*}} dy\int_{\Phi^{-1} (\Phi^*)}  \delta_{x_2^{**}} (x_2) \cdot p^I(y |x_1)\cdot  \delta_{x_1^*} (x_1) dx    \\
 &~~~~~~~~~~~~~~~~~~~~~~~~~~~+  \int_{\Y - N_{y^*}} dy\int_{\Phi^{-1} (\Phi^*)}\delta_{x_2^{*}} (x_2) \cdot p^I(y |x_1) \cdot \delta_{x_1^*} (x_1) dx \\
 &=  \int_{N_{y^*}} dy\int_{\Phi^{-1} (\Phi^*)}  \delta_{(x_1^*, x_2^{**})} (x_1, x_2) \cdot p^I(y |x_1) dx    \\
 &~~~~~~~~~~~~~~~~~~~~~~~~~~+  \int_{\Y - N_{y^*}} dy\int_{\Phi^{-1} (\Phi^*)}\delta_{(x_1^*, x_2^{*})}(x_1, x_2) \cdot p^I(y |x_1)  dx. 
 \end{align*}
Noting that $(x_1^*, x_2^{*})  \in \Phi^{-1} (\Phi^*)$ and $(x_1^*, x_2^{**})  \notin \Phi^{-1} (\Phi^*)$, we obtain 
 \begin{align*}
P_{X^{b}} ( \Phi^{-1} (\Phi^*))  &=  \int_{N_{y^*}} dy\int_{\Phi^{-1} (\Phi^*)}  \delta_{(x_1^*, x_2^{**})} (x_1, x_2) \cdot p^I(y |x_1) dx    \\
 &~~~~~~~~~~~~~~~~+  \int_{\Y - N_{y^*}} dy\int_{\Phi^{-1} (\Phi^*)}\delta_{(x_1^*, x_2^{*})}(x_1, x_2) \cdot p^I(y |x_1)  dx \\
 &=  \int_{N_{y^*}} dy \cdot 0     
 +  \int_{\Y - N_{y^*}} dy \cdot p^I(y |x_1^*) \\
 &=  \int_{\Y - N_{y^*}}  p^I(y |x_1^*)  dy  \neq 0.
 \end{align*}

Hence, we obtain  
 \begin{align*}
 P_{Y^{b} |\Phi(X^{b})} (N_{y^*} | \Phi^*) &= \frac{ P_{X^{b}, Y^{b}} ( \Phi^{-1} (\Phi^*) \times N_{y^*} ) }{P_{X^{b}} ( \Phi^{-1} (\Phi^*)) }  \\
 &= \frac{ 0}{\int_{\Y - N_{y^*}}  p^I(y |x_1^*)} = 0.
 \end{align*}
 Combing these results, we obtain 
 \begin{align*}
 P_{Y^{a} |\Phi(X^{a})} (N_{y^*} | \Phi^*) = 1 \neq 0 =P_{Y^{b} |\Phi(X^{b})} (N_{y^*} | \Phi^*) , 
 \end{align*}
 which contradicts the assumption 
 \begin{align*}
 P_{Y^{a}  |\Phi(X^{a})}=P_{Y^{b} |\Phi(X^{b})}.
 \end{align*}

 Because the continuity of $\Psi$  is trivial, we can conclude the proof.  \qed
 
 %The continuity of $\Psi^* $ is derived as follows:
 %take an open set $U \in \X_1$. Since $\Phi^*$ is continuous, 
 %$$(\Phi^*)^{-1} (U) = (\Psi^* \circ \Phi^{\X_1})^{-1} (U) = ( \Phi^{\X_1} )^{-1}\bigl( (\Psi^*)^{-1} (U)\bigr) = (\Psi^*)^{-1} (U) \times \X_2$$
% is open. Since the projection onto $\X_1$ is an open map \citep[Theorem 2 in Chapter 3]{kelley2008}, $(\Psi^*)^{-1} (U)$ is open, which concludes the continuity of $\Psi^*$. \qed

Finally, we prove Theorem \ref{theo:LSLsetting}.

\paragraph{\bf{Proof of Theorem  \ref{theo:LSLsetting}}} Take 
\begin{equation}\label{eq:min_IRM}
f^* \in \argmin \nolimits_{\Phi \in \I_{tr}^{C^0}, \\
w \in \W^{\C_0}}  \sum_{e \in \E_{tr}} \R^{e}(w \circ \Phi ).
\end{equation}
Then, by Lemma \ref{Lem:invaiance=compiose}, we can represent $f^*$ as $$f^* = w^* \circ \Phi^{\X_1}$$ for some continuous map $w^* : \X_1 \rightarrow \Y$\footnote{\st{By Lemma \ref{Lem:invaiance=compiose}, $f^*$ can be represented by $f^* = w^* \circ \Psi^* \circ \Phi^{\X_1}$ for $\Psi^*: \X_1 \rightarrow \H$ and $w^*: \H \rightarrow \Y$. Replacing $w^* \circ \Psi^*$ by $w^*$, we can obtain the desirable statement.}}. 
Let us prove that $w^* \circ \Phi^{\X_1} \in  \argmin_{f:\X \rightarrow \Y} \R^{o.o.d.} (f)$ by contradiction; assuming  that $ w^* \circ \Phi^{\X_1}  \notin \argmin_f \R^{o.o.d.} (f)$, we will derive  $w^* \circ \Phi^{\X_1} \notin \argmin \nolimits_{\Phi \in \I_{tr}^{C^0}, \\
w \in \W^{\C_0}}  \sum_{e \in \E_{tr}} \R^{e}(w \circ \Phi )$, which contradicts to  (\ref{eq:min_IRM}). We prove  %$w^* \circ \Phi^{\X_1} \notin \argmin \nolimits_{\Phi \in \I_{tr}^{C^0}, \\
%w \in \W^{\C_0}}  \sum_{e \in \E_{tr}} \R^{e}(w \circ \Phi )$ 
it by the following three steps. 

\paragraph{Step 1} First, we prove that there exist a training domain $e^{**} \in \E_{\text{tr}}$ and an open set $N_1 \subset \X_1$ which satisfy 
\begin{equation}\label{eq:step1}
w^*  (x_1) \neq  w^I (x_1)  \text{ for }  \forall x_1 \in N_1 
\end{equation}
with  $P_{X^{e^{**}}_1}(N_1) > 0$.
As $w^I \circ \Phi^{\X_1}$ minimizes the o.o.d. risk (Lemma \ref{Lem:nonlin_rojas}), we have
\begin{equation*}
 \R^{o.o.d.} (w^* \circ  \Phi^{\X_1}  ) >   \min_{f:\X \rightarrow \Y} \R^{o.o.d.} (f) =  \R^{o.o.d.}(w^I \circ \Phi^{\X_1}).
 \end{equation*}
Here, the first inequality is derived from the assumption of a proof by contradiction. Noting that $\R^{o.o.d.}$ is maximum of risk among $\{(X^e, Y^e)\}$, there exists $(X^{e^*}, Y^{e^*}) \in \{(X^e, Y^e)\}_{e \in \E}$ such that 
\begin{align}
     &\R^{e^*} (w^* \circ \Phi^{\X_1}) ~~ \Bigl( = \int  \| y-  w^* (x_1)\|^2 dP_{X^{e^*}_1, Y^{e^*}} \Bigr) \notag  \\
    &~~~~~~~~~~~~~~~~~~~~~~~~~~~~~~~ >   \R^{e^*} (w^I \circ \Phi^{\X_1}) ~~\Bigl( = \int \| y- w^I (x_1) \|^2  dP_{X^{e^*}_1, Y^{e^*}}  \Bigr) \label{ineq:contr_ood}
\end{align}
holds\footnote{Note that $e^*$ is not necessarily included in training domains $\E_{tr}$. The inequality (\ref{ineq:contr_ood}) for some training domain $e^{**} \in \E_{tr}$ are proved in Step 2 (eq. (\ref{ineq:w_not_min})). }. Since (\ref{ineq:contr_ood}) is rewritten as 
$$ \int \left\{ \| y- w^* (x_1) \|^2 -  \| y- w^I  (x_1) \|^2 \right\}dP_{X^{e^*}_1, Y^{e^*}} > 0,$$ 
we can see that 
$$\| y^* - w^* (x^*_1) \|^2 -  \| y^* - w^I  (x^*_1) \|^2 >0$$ for some $( x^*_1, y^*) \in \X_1 \times \Y$. Since $w^*$ and $w^I$ are continuous, taking sufficiently small $\e >0$, we have 
\begin{equation}\label{eq:square_posi}
\| y^* - w^*  (x_1) \|^2 -  \| y^* - w^I (x_1) \|^2 >0  \text{ for } \forall x_1 \in N_{x_1^*}^{\e},
\end{equation}
where  $N_{x_1^*}^{\e}$  is the $\e$-ball centered at $x_1^*$. Here, the continuity of $w^I$ is derived from  Condition (iv) in Theorem 1. Moreover, (\ref{eq:square_posi}) leads us to the statement 
 \begin{center}
$ w^*  (x_1)\neq  w^I (x_1)$  for $\forall x_1 \in  N_{x_1^*}^{\e}$.
\end{center}
By the condition (ii),  $ N_{x_1^*}^{\e}  \bigcap \text{supp}(P_{X^{e^{**}}_1}) \neq \emptyset$ for some $e^{**} \in \E_{\text{tr}}$. Take 
\begin{align*}
    &x^{**}_1 \in N_{x_1^*}^{\e} \bigcap \text{supp}(P_{X^{e^{**}}_1}) \stackrel{\text{def of supp}}{=\hspace{-1mm}=\hspace{-1mm}=} \\
    &~~~~~ N_{x_1^*}^{\e} \bigcap \overline{ \left\{x_1 \in \X_1 \left|  N_{x_1}: \st{\text{open neighborhood around }x_1}  \Rightarrow  (P_{X^{e^{**}}_1})(N_{x_1}) > 0 \right. \right\}} \neq \emptyset.
\end{align*}
    Replacing $x^{**}_1$, if necessary, we may assume that 
\begin{equation}\label{assump:replacement}
 x^{**}_1 \in  N_{x_1^*}^{\e} \bigcap \left\{x_1 \in \X_1 \left|  N_{x_1}: \st{\text{open neighborhood around }x_1}  \Rightarrow  (P_{X^{e^{**}}_1})(N_{x_1}) > 0 \right. \right\}.
%\footnote{
% For $x^{**}_1 \in  N_{x_1^*}^{\e} \bigcap \text{supp}(P_{X^{e^{**}}_1})$, set  an open neighborhood $N \subset  N_{x_1^*}^{\e}$ with  $x^{**}_1 \in N$. Since $x_1^{**} \in  \text{supp}(P_{X^{e^{**}}_1})$,  we can see that  
%$$N \bigcap \left\{x_1 \in \X_1 \left|  N_{x_1}: \text{open neighborhood with }x_1 \in N_{x_1}  \Rightarrow (P_{X^{e^{**}}_1})(N_{x_1}) > 0 \right. \right\} \neq \emptyset$$ by the definition of closure. Replacing $x_1^{**}$ by an element in the above non-empty intersection, if necessary, we can justify the assumption (\ref{assump:replacement}). }
\end{equation}
Take an open set $N_{1} \subset  N_{x_1^*}^{\e}$ which includes $x^{**}_1$. Then,  we have
\begin{equation}\label{state:not_corresp_nonzeomeas}
 w^*  (x_1) \neq  w^I (x_1)  \text{ for }  \forall x_1 \in N_{1}.
\end{equation}
 Observing that 
$$ x^{**}_1 \in  \left\{x_1 \in \X_1 \left|  N_{x_1}: \text{open neighborhood with }x_1 \in N_{x_1}  \Rightarrow (P_{X^{e^{**}}})(N_{x_1}) > 0 \right. \right\}, $$
we have $P_{X^{e^{**}}_1}(N_{1}) > 0$. It concludes the proof of Step 1. 
\paragraph{Step 2} Next, we prove the inequality 
\begin{equation}\label{ineq:step2}
      \sum_{e \in \E_{tr}} \R^{e}(w^* \circ \Phi^{\X_1} ) > \sum_{e \in \E_{tr}} \R^{e}(w^I \circ \Phi^{\X_1} ). 
\end{equation}
To derive the inequality, note that 
\begin{equation*}
    \hat{w} \in \argmin_w \R^{e} (w \circ \Phi^{\X_1}) \Longleftrightarrow \hat{w}(x_1) = w^I(x_1) ~~P_{X^e_1}-\text{a.e}., %\footnote{While the inverse implication also holds, it is not utilized in our proof.}
\end{equation*}
or equivalently, 
\begin{equation}\label{state:min_is_condi_exp_ae}
    \hat{w} \in \argmin_w \R^{e} (w \circ \Phi^{\X_1}) \Longleftrightarrow P_{X^e_1} (\left\{x_1 \in \X_1  \left| \hat{w} (x_1) \neq w^I (x_1 )\right. \right\}) =0 
\end{equation}
holds for any $e \in \E$ \citep[Example 2.6]{andreas2008}. Taking the contraposition of the implication from the left to right propositions in (\ref{state:min_is_condi_exp_ae}), we have
\begin{equation}\label{impli_notcondi}
   \hat{w}\text{ satisfies }P_{X^e_1} (\left\{x_1 \in \X_1  \left|\hat{w}(x_1) \neq w^I (x_1 )\right. \right\}) > 0  \Rightarrow  \hat{w} \notin \argmin_w \R^{e} (w \circ \Phi^{\X_1}) .
\end{equation}
From (\ref{eq:step1}), we have the inequality 
\begin{equation}\label{ineq:meas_nonzeo}
   P_{X^{e^{**}}_1} (\left\{x_1 \in \X_1  \left| w^* (x_1) \neq w^I (x_1 )\right. \right\}) > P_{X^{e^{**}}_1} (N_{1}) > 0 
\end{equation}
 for  some $e^{**} \in \E_{tr}$ and an open set $N_1 \subset \X_1$. (\ref{impli_notcondi}) and (\ref{ineq:meas_nonzeo}) lead us to statement ${w^*} \notin \argmin_w \R^{e^{**}} (w \circ \Phi^{\X_1})$, and hence, we have the inequality  
\begin{equation}\label{ineq:w_not_min}
\R^{e^{**}} (w^* \circ \Phi^{\X_1}) > \min_w \R^{e^{**}} (w \circ \Phi^{\X_1}) = \R^{e^{**}} (w^I \circ \Phi^{\X_1}).
\end{equation}
Moreover, since the conditional expectation $w^I$ minimizes the risk, we have
\begin{equation}\label{ineq:w_geq_condi}
\R^e(w^* \circ \Phi^{\X_1}) \geq \R^e(w^I \circ \Phi^{\X_1})
\end{equation}
 for any $e \in \E$. (\ref{ineq:w_not_min}) and (\ref{ineq:w_geq_condi}) lead us to the inequality
\begin{align*}
    \sum_{e \in \E_{tr}} \R^{e}(w^* \circ \Phi^{\X_1} ) &= \R^{e^{**}}(w^* \circ \Phi^{\X_1} ) +\sum_{e \in \E_{tr} - \{ e^{**}\}} \R^{e}(w^* \circ \Phi^{\X_1} )  \\
    & \stackrel{(\ref{ineq:w_not_min})}{>}  \R^{e^{**}}(w^I \circ \Phi^{\X_1} ) +\sum_{e \in \E_{tr} - \{ e^{**}\}} \R^{e}(w^* \circ \Phi^{\X_1} )\\
    & \stackrel{(\ref{ineq:w_geq_condi})}{\geq} \R^{e^{**}}(w^I \circ \Phi^{\X_1} ) +\sum_{e \in \E_{tr} - \{ e^{**}\}} \R^{e}(w^I \circ \Phi^{\X_1} ) = \sum_{e \in \E_{tr}} \R^{e}(w^I \circ \Phi^{\X_1} ).
\end{align*} 
%Since $w^I \in \W^{\C_0}$ and $\Phi^{\X_1} \in \I_{tr}^{\C_0}$, the inequality contradicts to the assumption (\ref{eq:min_IRM}), which concludes the proof. 

\paragraph{Step 3} Finally, we prove $w^* \circ \Phi^{\X_1} \notin \argmin \nolimits_{\Phi \in \I_{tr}^{C^0}, \\
w \in \W^{\C_0}}  \sum_{e \in \E_{tr}} \R^{e}(w \circ \Phi )$, which contradicts to  (\ref{eq:min_IRM}). By the inequality (\ref{ineq:step2}) proved in Step 2, it suffices to prove that there exist $\Phi^\dagger \in \I_{tr}^{\C_0}$ and $w^\dagger \in \W^{\C_0}$ such that $ w^I \circ \Phi^{\X_1} = w^\dagger \circ \Phi^\dagger$. Define  $\Phi^{\dagger} =  \Psi^{\dagger}  \circ \Phi^{\X_1}$ where the embedding  $\Psi^{\dagger} : \X_1~(= \bR^{d_1}) \rightarrow \H~(= \bR^{d_\H})$ is defined by 
$$  \bR^{d_1} \ni \begin{pmatrix}
 x^1 \\ x^2 \\ \vdots \\ x^d 
\end{pmatrix}  \stackrel{\Psi^{\dagger}}{\longmapsto} \begin{pmatrix}
x^1 \\ x^2 \\ \vdots \\ x^d \\ 0 \\ \vdots  \\ 0
\end{pmatrix} \in \bR^{d_\H}. $$ Here, we can define the embedding $\Psi^{\dagger}$ since \st{$d_1 \leq d_\H$ }(Condition (iii)).
Noting that  $P_{Y^e | \Phi^\dagger (X^e)} = P_{Y^e | \Phi^{\X_1} (X^e)} = P_{Y^I | X_1^I}$ for any $e \in \E$, we can see that $\Phi^{\dagger} \in \I^{\C_0}_{tr}$. Defining 
 $$ \bR^{d_\H} \ni \begin{pmatrix}
 x^1 \\ x^2 \\ \vdots \\ x^d \\ x^{d+1} \\ \vdots \\ x^{h}
\end{pmatrix}    \stackrel{w^{\dagger}}{\longmapsto} \bE[Y^I | X_1^I = \begin{pmatrix}
 x^1 \\ x^2 \\ \vdots \\ x^d 
\end{pmatrix} 
] \in \Y , $$
 we can see that $ w^I \circ \Phi^{\X_1} = w^\dagger \circ \Phi^\dagger$. Observing $w^\dagger \in \W^{\C_0}$ by Condition (iv), we can concludes $w^* \circ \Phi^{\X_1} \notin \argmin \nolimits_{\Phi \in \I_{tr}^{C^0}, \\
w \in \W^{\C_0}}  \sum_{e \in \E_{tr}} \R^{e}(w \circ \Phi )$,  which contradicts to  (\ref{eq:min_IRM}). 
\qed

%-------------------------------------------------------------------------------------------------------------------------------------------------------------------
\subsection{Proof of Theorem \ref{theo:CEsetting}}
%-------------------------------------------------------------------------------------------------------------------------------------------------------------------

We prepare two lemmas.
\begin{lemm}\label{Lem:nonlin_rojas_CE}
Let $p^I: \X_1 \rightarrow \P_{\Y}$ be the conditional p.d.f. of $P_{Y^I | X_1^I}$; namely, 
$$ \bigl( p^I (x) \bigr)_i := p^I (i | x). $$
Then, 
$$p^I \circ \Phi^{\X_1} \in \argmin_{f: \X \rightarrow \P_{\Y} } \R^{o.o.d.}(f).$$
\end{lemm}

\begin{lemm}\label{Lem:invaiance=compiose_CE}
Any $\Phi \in \I_{tr}^{C^0}$ is represented as 
$$  \Phi = \Psi \circ \Phi^{\X_1} $$
for some continuous map $\Psi : \X_1 \rightarrow \H$.
\end{lemm}

\paragraph{\bf{Proof of Lemma  \ref{Lem:nonlin_rojas_CE}}}
The proof is essentially the same as the ones for Lemma \ref{Lem:nonlin_rojas}; hence, we omit the proof. \qed

\paragraph{\bf{Proof of Lemma \ref{Lem:invaiance=compiose_CE}}}
First, we prove that $\Phi \in \I_{tr}^{C^0}$ can be represented as 
\begin{equation}\label{min_compose3}
\Phi = \Psi \circ \Phi^{\X_1}
\end{equation}
by some map $\Psi : \X_1 \rightarrow \X_1$, which is not restricted to a continuous map.
We prove this statement by contradiction in the same manner as the proof in Lemma \ref{Lem:invaiance=compiose}. Take $\Phi \in \I_{tr}^{C^0}$. Then, there exist $x_1^* \in \X_1$, $x_2^*, x_2^{**} \in \X_2$ such that 
$$  \Phi (x_1^*, x_2^*) \neq \Phi(x_1^*, x_2^{**}).$$ Fix $y^* \in \Y$ with $p^I(y^*|x_1^*) > 0$. 
Define two maps $g^i: \Y \rightarrow \X_2$ ($i=1,2$) by 
%\vspace{5mm}
 \begin{center}
  $ g^1( y) = \left\{
\begin{array}{ll}
 x_2^* & (y =y^*)\\
x_2^{**} & (\text{ else })
\end{array}
\right. $ \textcolor{white}{aaaaaa} 
$g^2( y) = \left\{
\begin{array}{ll}
  x_2^{**} & (y =y^*)\\
x_2^{*} & (\text{ else })
\end{array}
\right. $
 \end{center}
Take two distributions $(X^{a}, Y^{a} ), (X^{b}, Y^{b} ) \in \{ (X^e, Y^e) \}_{e \in \E}$   such that their distributions $P_{X^{a}, Y^{a}}$ and   $P_{X^{b}, Y^{b}}$ coincide with 
 \begin{align*}
P_{X^{a}, Y^{a}} = P_{X_2^{a} | Y^{a}} \otimes  P_{Y^I | X_1^I }
\otimes 
P_{X_1 }, ~~~~~
P_{X^{b}, Y^{b}} = P_{X_2^{b}  |Y^{b} } \otimes  P_{Y^I | X_1^I }
\otimes 
P_{X_1 }.
\end{align*}
Here
\begin{itemize}
\item  $P_{X_1 }$ is a  distribution on $\X_1$ where its p.d.f. coincides with  a delta function $ \delta_{x_1^*} (x_1)$ on $x_1^*$,
\item the conditional p.d.f.s of $P_{X_2^{a} | Y^{a}}(\cdot | y)$ and $P_{X_2^{b} | Y^{b}}(\cdot | y)$    coincide with $\delta_{g^1(y)} (x_2)$ and  $ \delta_{g^{2}(y)} (x_2) $ respectively.
\end{itemize} 
Since $\Phi \in \I^{\C_0} = \I^{\C_0}_{tr}$ (Condition (i)) and $(X^{a}, Y^{a} ), (X^{b}, Y^{b} ) \in \{ (X^e, Y^e) \}_{e \in \E}$,
 \begin{equation}\label{eq:invairnace_CE}
   P_{Y^{a}|\Phi(X^{a})} (\{ y^* \} | \Phi^* ) =   P_{Y^{b}|\Phi(X^{b})} (\{ y^* \} | \Phi^* ) ,
 \end{equation}
 where  $\Phi^*: = \Phi(x_1^*, x_2^*)$.
Let us compute $P_{Y^{a}|\Phi(X^{a})} (\{ y^* \} | \Phi^* )$ and $P_{Y^{b}|\Phi(X^{b})} (\{ y^* \} | \Phi^* )$, respectively. Same as the proof in Lemma \ref{Lem:invaiance=compiose}, we have
 the two equalities $$ P_{\Phi(X^{a}), Y^{a}} ( \{ \Phi^* \} \times \{ y^* \} )  = p^I(y^* |x_1^*) \text{ and }P_{\Phi(X^{a})} (  \{ \Phi^* \} ) =  p^I(y^* |x_1^*), $$  which lead us to the equality 
 \begin{align*}
 P_{Y^{a}|\Phi(X^{a})} ( \{ y^* \} | \Phi^* )  &= \frac{ P_{\Phi(X^{a}), Y^{a}} ( \{ \Phi^* \} \times \{ y^* \} )}{P_{\Phi(X^{a})} (  \{ \Phi^* \} )} \\
 &= \frac{p^I(y^* |x_1^*) }{p^I(y^* |x_1^*)} = 1.
 \end{align*}
 Similarly, we  have 
 $$ P_{Y^{b}|\Phi(X^{b})} ( \{ y^* \} | \Phi^* )  =0 \text{ and }P_{\Phi(X^{b})} (  \{ \Phi^* \} ) =   \sum_{y \in \Y - \{ y^* \}} p^I(y |x_1^*) \neq 0, $$
 which lead us to the equality
 \begin{align*}
 P_{Y^{b}|\Phi(X^{b})} ( \{ y^* \} | \Phi^* )  &= \frac{ P_{\Phi(X^{b}), Y^{b}} ( \{ \Phi^* \} \times \{ y^* \} )}{P_{\Phi(X^{b})} (  \{ \Phi^* \} )} \\
 &= \frac{0}{\sum_{y \in \Y - \{ y^* \}} p^I(y |x_1^*)} = 0.
 \end{align*}
  Here, $\sum_{y \in \Y - \{ y^* \}} p^I(y |x_1^*) \neq 0$ is derived by Condition (v).
 Combing these results, we obtain 
 \begin{align*}
  P_{Y^{a}|\Phi(X^{a})} ( \{ y^* \} | \Phi^* ) = 1 \neq 0 = P_{Y^{b}|\Phi(X^{b})} ( \{ y^* \} | \Phi^* ) , 
 \end{align*}
 which contradicts the assumption 
 \begin{align*}
 P_{Y^{a}  |\Phi(X^{a})}=  P_{Y^{b}  |\Phi(X^{b})}.
 \end{align*}
 Because the continuity of $\Psi$  is trivial, we can conclude the proof. 
 \qed
 
 %The continuity of $\Psi^* $ is same as one in the proof for  Lemma \ref{Lem:invaiance=compiose}, and hence, we omit. \qed

%\paragraph{\bf{Proofs  of Lemmas \ref{Lem:minIRM_CE} and  \ref{Lem:charactrizeIRM_CE}}} 

%Proofs of Lemmas \ref{Lem:minIRM_CE} and \ref{Lem:charactrizeIRM_CE} are same as ones of Lemmas \ref{Lem:minIRM} and \ref{Lem:charactrizeIRM}, and hence, we omit the proof. \qed

\paragraph{\bf{Proof of Theorem \ref{theo:CEsetting}}}
This is essentially the same as  the one for Theorem 1, and hence, we omit the proof. \qed

%-------------------------------------------------------------------------------------------------------------------------------------------------------------------
%\subsection{Proof of Theorem 3} 
%-------------------------------------------------------------------------------------------------------------------------------------------------------------------

%-------------------------------------------------------------------------------------------------------------------------------------------------------------------
\section{Conclusions}\label{sec:conclusion}
In this paper, we have proved that a solution for the bi-leveled optimization problem (\ref{eq:IRM}) also minimizes o.o.d.~risk (\ref{eq:oodMinimization})  under four conditions in regression and classification cases, assuming that distributions on domains are the ones proposed  in \citet{rojas-carulla2018a} and that models run all continuous functions.  Particularly, we have provided a sufficient condition on the training domains $\E_{\text{tr}}$ and the dimension of the feature space $\H$ for the optimization problem (\ref{eq:IRM}) to minimize the  o.o.d. risk. %: invariances $\I_{tr}$ captured by training domains $\E_{tr}$ corresponds to ones $\I$ by all domains $\E$. \citet{arjovsky2020} also pointed out that the condition $\I = \I_{tr}$ facilitates o.o.d. generalization,  and our result gives rigorous support for their discussion. %This result provides theoretical support for the discussion by \cite{arjovsky2020}. %Our results  provide a solid theoretical foundation for using the bi-leveled optimization problem (\ref{eq:IRM}) to the o.o.d.~generalization problems.

%There are still many open problems left.  
Several challenges still exist. The first problem  is the theoretical analysis of the optimization method for (\ref{eq:IRM}). To solve the challenging optimization problem (\ref{eq:IRM}), various optimization techniques have been proposed \citep{arjovsky2020, lin2022, zhou2022}, and there has been little discussion about their effectiveness. For example, while \citet{arjovsky2020} optimized (\ref{eq:IRM}) by minimizing 
$$\sum_{e \in \E_{tr}} \R^e (\Phi) + \lambda \cdot  \| \nabla_{w | w = 1.0}  \R^e (w \cdot \Phi) \|^2, $$
 their effectiveness was evaluated only under specific SEMs \citep{rosenfeld2021}. Thus, it is important to investigate this analysis under a more general case.

Second, we should evaluate the o.o.d.~performance of  the bi-leveled optimization problem (\ref{eq:IRM}) under the case where the  conditions in Theorems \ref{theo:LSLsetting} and \ref{theo:CEsetting} are violated. Particularly, as noted in Section \ref{sec:Main}, condition $\I_{tr} = \I$ does not generally hold. In such cases, for $(\Phi^*, w^*) \in \argmin \nolimits_{\Phi \in \I_{tr}^{\C^0}, \\
w \in \W^{\C_0}}  \sum_{e \in \E_{tr}} \R^{e}(w \circ \Phi ),$
$$ \R^{o.o.d} (w^* \circ \Phi^*) - \min\R^{o.o.d} (f)$$ is not necessarily $0$.  The quantitative evaluation of the difference is crucial for future work. 

\st{Thirdly, we should investigate the feasibility of the condition $\I_{tr} = \I$, which is known to be an important and unsolved problem  shared by all invariance-based methods \citep{arjovsky2020, peters2016, toyota2022}. As Condition (i) in our main results, all methods based on invariances implicitly or explicitly assume that invariances among training domains correspond to ones among all domains. As discussed in Section \ref{sec:Main},  some sufficient conditions under a simple linear SEM setting have been found \citep{peters2016, arjovsky2020}, but general theoretical results have not yet been established. This is also among our unsolved problems, and should be provided in further work.}
%As noted in Section \ref{sec:Main}, on domain setting (\ref{theo:assump1}), both sufficient and necessary conditions for $\I_{tr} = \I$ are not yet been revealed. 
%This should be provided in further work.}

\st{Finally, extending our results to general domain sets beyond the case by \citet{rojas-carulla2018a} is an important topic for future work. Invariant Risk Minimization (IRM) estimates the feature map $\Phi$ that has the same conditional distribution $P_{Y^e|\Phi(X^e)}$ among all domains $e \in \E$; in other words, IRM framework assumes that a domain set $\E$ has a feature map $\Phi$ such that $P_{Y^e|\Phi(X^e)}$ are equal among all domains. Among domain sets that satisfy the property, the domain set by \citet{rojas-carulla2018a} is the simplest one; the projection $\Phi^{\X_1}$ induces the same conditional independence $P_{Y^e|\Phi^{\X_1}(X^e)}$. In some cases, a map that induces the same conditional distribution is a more complex function than the projection $\Phi^{\X_1}$, so the relation between (\ref{eq:IRM}) and the o.o.d.~risk on such general domains beyond the case by \citet{rojas-carulla2018a} is should be investigated.}
%-------------------------------------------------------------------------------------------------------------------------------------------------------------------

%%%==============================================
\section*{Acknowledgements}
%==============================================
We thank Dr.~Yano in the Institute of Statistical Mathematics for valuable discussions. The research was supported by  Grant-in-Aid for JSPS Fellows 20J21396, Grant-in-Aid for Research Activity Start-up 23K19966, JST CREST JPMJCR2015, and JSPS Grant-in-Aid for Transformative Research Areas (A) 22H05106.

\bibliography{IRM_withoutSEM}
\bibliographystyle{tmlr}

\end{document}